\documentclass[10pt, a4paper]{article}

\usepackage[final]{lrec-coling2024}
\usepackage{inconsolata}
\usepackage{amssymb}
\usepackage{dsfont}
\usepackage{amsmath}
\usepackage{booktabs}
\usepackage{multirow}
\usepackage[normalem]{ulem}
\useunder{\uline}{\ul}{}
\usepackage{amssymb}
\usepackage{bbm}
\usepackage{graphicx}
\usepackage{colortbl}
\usepackage{amsthm}
\usepackage{mathrsfs}
\usepackage[linesnumbered,ruled,vlined]{algorithm2e}

\title{KEHRL: Learning Knowledge-Enhanced Language Representations with Hierarchical Reinforcement Learning}

\name{Dongyang Li$^{1,2}$,  {\bf \large Taolin Zhang$^{2}$,}  {\bf \large Longtao Huang$^2$,} \\ {\bf \large Chengyu Wang$^{2}$,}  {\bf \large Xiaofeng He$^{1}$,\thanks{Work done when Dongyang Li was doing an internship at Alibaba Group. Dongyang Li and Taolin Zhang contributed equally to this work. Correspondence to Chengyu Wang and Xiaofeng He.}}  {\bf \large Hui Xue$^2$}}

\address{$^1$School of Computer Science and Technology, East China Normal University \\$^2$Alibaba Group \\
         dongyangli0612@gmail.com, hexf@cs.ecnu.edu.cn\\
         \{zhangtaolin.ztl, kaiyang.hlt, chengyu.wcy, hui.xueh\}@alibaba-inc.com\\}

\abstract{
Knowledge-enhanced pre-trained language models (KEPLMs) leverage relation triples from knowledge graphs (KGs) and integrate these external data sources into language models via self-supervised learning.
Previous works treat knowledge enhancement as two independent operations, i.e., knowledge injection and knowledge integration.
In this paper, we propose to learn \textbf{K}nowledge-\textbf{E}nhanced language representations with \textbf{H}ierarchical \textbf{R}einforcement \textbf{L}earning (KEHRL), which jointly addresses the problems of detecting positions for knowledge injection and integrating external knowledge into the model in order to avoid injecting inaccurate or irrelevant knowledge.
Specifically, a high-level reinforcement learning (RL) agent utilizes both internal and prior knowledge to iteratively detect essential positions in texts for knowledge injection, which filters out less meaningful entities to avoid diverting the knowledge learning direction.
Once the entity positions are selected, a relevant triple filtration module is triggered to perform low-level RL to dynamically refine the triples associated with polysemic entities through binary-valued actions.
Experiments validate KEHRL's effectiveness in probing factual knowledge and enhancing the model's performance on various natural language understanding tasks.
 \\ \newline  }

\begin{document}

\maketitleabstract

\section{Introduction}
General pre-trained language models (PLMs) \cite{DBLP:conf/nips/00040WWLWGZH19,DBLP:conf/emnlp/LiZHWYL20,DBLP:conf/icml/Bao0WW0L0GP0H20} are pre-trained on various sources \cite{DBLP:conf/coling/MaCSLWH20,DBLP:conf/cikm/WuH19a,DBLP:conf/icml/GuuLTPC20} and fine-tuned with specific data for diverse tasks, such as Information Extraction \cite{DBLP:conf/acl/LeeLDPSHAWFP22,DBLP:conf/acl/QinLT00JHS020,DBLP:conf/acl/MaGLZHZ20}, Natural Language Inference \cite{DBLP:conf/acl/QiWDC22,DBLP:conf/emnlp/SahaNB20}, and Question Answering \cite{DBLP:conf/acl/ZhangZY000C22,DBLP:conf/acl/HeoKCZ22,DBLP:conf/acl/PappasA20,DBLP:conf/acl/0002SLHCG20}.

To enhance context-aware representations, PLMs are equipped with additional knowledge collected from external resources in the forms of structured data such as relation triples from knowledge graphs (KGs) \cite{DBLP:journals/aiopen/SuHZLLLZS21} and unstructured description texts related to entities \cite{DBLP:conf/acl/00020FYWX0022}. This type of PLMs is often called knowledge-enhanced pre-trained language models (KEPLMs).
Meanwhile, the recently emerged large language models \cite{DBLP:journals/corr/abs-2305-03268,DBLP:journals/corr/abs-2302-12813} also need external knowledge such as parametric knowledge \cite{DBLP:journals/corr/abs-2305-04757} and retrieved knowledge \cite{retrieval_llm} to augment themselves to alleviate hallucination \cite{DBLP:journals/corr/abs-2305-13669,DBLP:journals/corr/abs-2306-05212}.

\begin{figure*}[!htb]
\centering
\includegraphics[width=0.9\textwidth]{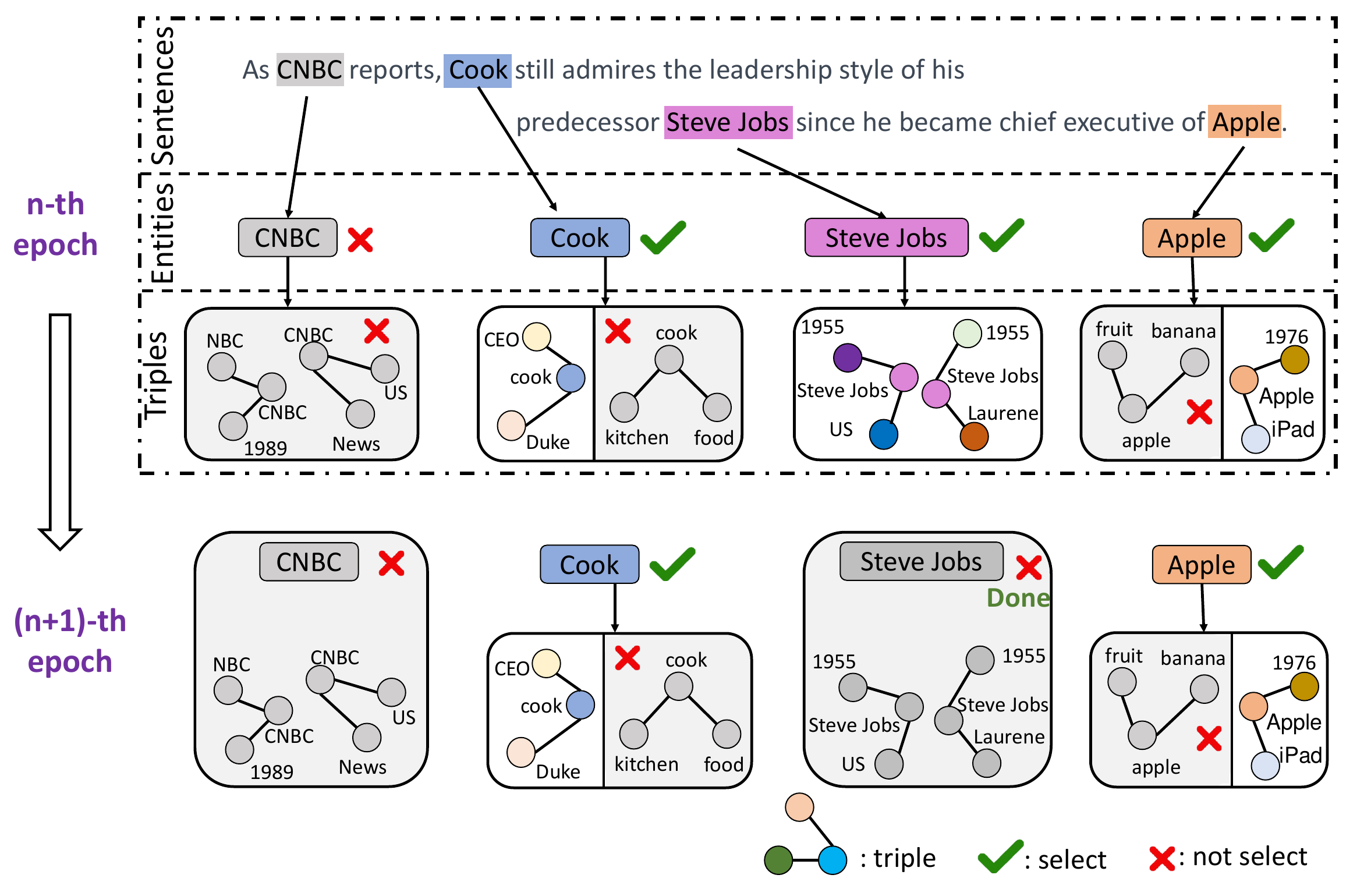} 
\caption{The dynamic selection process of entities and relation triples during pre-training.}
\label{motivation}
\end{figure*}

According to previous research, KEPLMs \cite{DBLP:journals/tacl/WangGZZLLT21,DBLP:conf/acl/ZhangC0QYH20,DBLP:conf/aaai/Zhang0HQTH022} generally consist of two important modules, namely, knowledge injection and knowledge integration.
(1) \textbf{Knowledge Injection}: it preprocesses the pre-training corpus into token-level input and chooses vital token positions (e.g., entities), preparing to inject the retrieved relevant relation triples from KGs at these positions.
Recent research \cite{DBLP:conf/acl/ZhangHLJSL19,DBLP:conf/emnlp/PetersNLSJSS19,DBLP:conf/dasfaa/LiuZGJ22} injects various types of knowledge into the positions of all entities indiscriminately.
However, frequent and commonly used entities have already been learned sufficiently by PLMs, causing the redundant learning phenomenon and further inducing knowledge noise \cite{DBLP:conf/ijcai/ZhangDCCZZC21,DBLP:conf/aaai/Zhang0HQTH022}.
(2) \textbf{Knowledge Integration}: it aggregates the retrieved relation triples into the context-aware entity representations output by a plain PLM and learns new knowledge-enhanced representations to produce the final KEPLM.
Existing works \cite{DBLP:conf/acl/ZhangHLJSL19,DBLP:conf/emnlp/LinCCR19,DBLP:conf/emnlp/PetersNLSJSS19} generally allocate different attention weights to each entity's triples to remove redundant knowledge.
However, the soft distribution of the attention mechanism is not deterministic aggregation and inevitably introduces inaccurate knowledge such as irrelevant ambiguity triples \cite{DBLP:conf/emnlp/PetersNLSJSS19,DBLP:conf/acl/ZhangHLJSL19}.
As shown in Figure \ref{motivation}, since ``CNBC'' has a relatively weak contribution to the meaning of the whole sentence, we should attach less emphasis to it to avert error propagation to subsequent procedures. ``Cook'' and ``Apple'' are polysemic entities, and their irrelevant triples should be filtered out to avoid inaccurate knowledge.
The well-learned entity ``Steve Jobs'' at the $n$-th epoch should not participate in the subsequent $(n+1)$-th epoch knowledge enhancement process to prevent duplicate learning \cite{DBLP:conf/ijcai/ZhangDCCZZC21,DBLP:conf/aaai/Zhang0HQTH022}.

To tackle the problems mentioned above, we propose a new \textbf{K}nowledge-\textbf{E}nhanced language representation learning framework with \textbf{H}ierarchical \textbf{R}einforcement \textbf{L}earning (KEHRL) process to alleviate the error propagation problem, which jointly learns the positions of entities for knowledge injection and leverages relevant candidate relation triples dynamically at different levels for knowledge pre-training. Two new techniques are proposed and summarized below.
\textbf{(1) Reinforced Entity Position Detection} combines the sentence's current representations and the prior knowledge as the state, leveraging high-level RL \cite{DBLP:books/lib/SuttonB98} to detect the essential entity positions
using the entity reward function derived based on the masked language modeling (MLM) task~\cite{DBLP:conf/naacl/DevlinCLT19}. 
The final entity positions guide the model toward the prospective direction.
When high-level RL's actions regarding entity positions are determined, low-level relevant triple selection will be triggered.
\textbf{(2) Reinforced Triple Semantic Refinement} utilizes low-level RL to choose semantically valid relation triples of polysemic entities with binary-valued actions. We dynamically prune inaccurate and ambiguous relation triples according to the current state. The MLM task's token accuracy reward guides the model to adjust itself to calibrate the learning bias from irrelevant relation triples.

\begin{table*}[tb]
\centering
\begin{tabular}{ccccc}
\toprule
Tasks                                       &  \begin{tabular}[c]{@{}c@{}}Named Entity\\ Recognition\end{tabular} & \begin{tabular}[c]{@{}c@{}}Relation\\ Extraction\end{tabular} & \begin{tabular}[c]{@{}c@{}}Sentiment\\ Analysis\end{tabular} & \begin{tabular}[c]{@{}c@{}}Information\\ Retrieval\end{tabular} \\ \midrule
Datasets                                    & ACE2005                 & SemEval             & SST-2              & MARCO DOC DEV         \\ \midrule
{$\quad$ Types $\downarrow$ $\;$ Metrics $\rightarrow$} & F1                      & F1                  & ACC                & MRR@100               \\ \midrule
No Entity                                         & 83.4                    & 91.7                & 93.5               & 39.8                  \\
All Entities                                         & 82.5                    & 89.6                & 92.5               & 38.3                  \\
Long-tail Entities                                 & 86.1                    & 95.2                & 95.8               & 41.1                  \\
High-frequency Entities                              & 84.8                    & 93.9                & 94.0               & 40.4                  \\ \bottomrule
\end{tabular}
\caption{The performance of different entity selection types on four different tasks.}
\label{data_analysis_entity}
\end{table*}

\section{Pre-training Data Analysis}
\label{appendix_pretrainig_data}
\noindent
\textbf{Selection of Entities}
We compare four different types of entity selection strategies for knowledge injection to observe changes in KEPLMs' performance, including (1) no entities selected, (2) all entities, (3) long-tail entities only, and (4) high-frequency entities only.
We utilize BERT-base~\cite{DBLP:conf/naacl/DevlinCLT19} as the backbone to evaluate the performance.
As Table \ref{data_analysis_entity} shows, we observe that knowledge injection into long-tail entities outperforms the high-frequency setting slightly, indicating that the models have already learned the factual knowledge well for these entities, where the relevant knowledge from KGs should be treated as redundant~\cite{DBLP:conf/ijcai/ZhangDCCZZC21,DBLP:conf/aaai/Zhang0HQTH022}.
The knowledge injection setting of all entities has the lowest scores compared to the others.
We suggest that the reason is that not all entities are helpful and different entities play different roles during the pre-training process for the enhancement of contextual semantics.

\begin{figure}[tb]
\centering
\includegraphics[width=0.95\columnwidth]{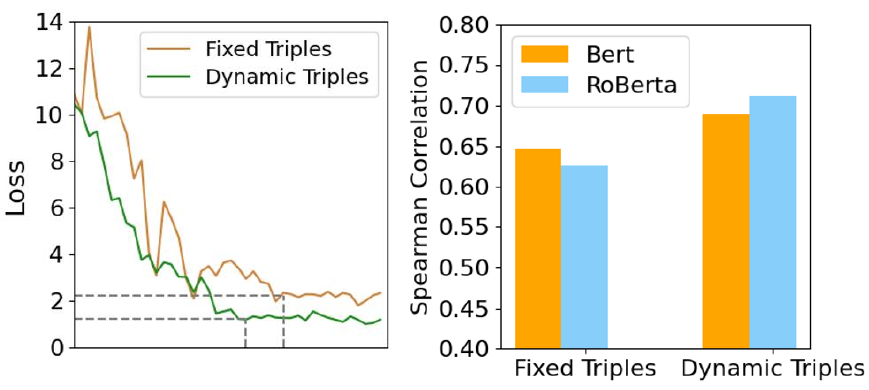}
\caption{The comparison of different triple injection operations.}
\label{data_analysis}
\end{figure}

\noindent
\textbf{Knowledge Injection from Relation Triples}
Since PLMs have incorporated knowledge into model parameters
\cite{DBLP:journals/aiopen/SuHZLLLZS21,DBLP:conf/coling/SunSQGHHZ20} through self-supervised learning, the remaining entities that are most difficult for the model to understand are polysemic entities \cite{DBLP:journals/tkde/ZhangLYFCX23}, such as the ambiguous semantics of ``apple'' regarding Apple Inc. or a kind of fruit.
Sentences injected with incorrect relation triples may
divert them from their correct meanings~\cite{DBLP:conf/aaai/LiuZ0WJD020}.
For example, when the important entity ``Apple Inc.'' occurs in the pre-training sentence, it would be harmful if a relation triple about the apple as a fruit is injected, such as ``$<$apple, subfamily\_neighbour, pear$>$''.
We analyze two relation triple injection operations, including fixed triple injection and dynamic triple injection.
Fixed triple injection leverages all correlated relation triples without any filtering techniques.
Dynamic triple injection absorbs relation triples with attention-weighted values learned by self-supervised knowledge pre-training tasks automatically.
We evaluate the average Spearman correlation score between the triple-injected training sentences and the original sentences.
From Figure \ref{data_analysis}, we observe that dynamic triple injection has higher Spearman scores than fixed triple injection, validating that the dynamically triple-injected sentences contain less inaccurate semantics.

\section{Model Architecture}
\begin{figure*}[tb]
\centering
\includegraphics[width=0.95\textwidth]{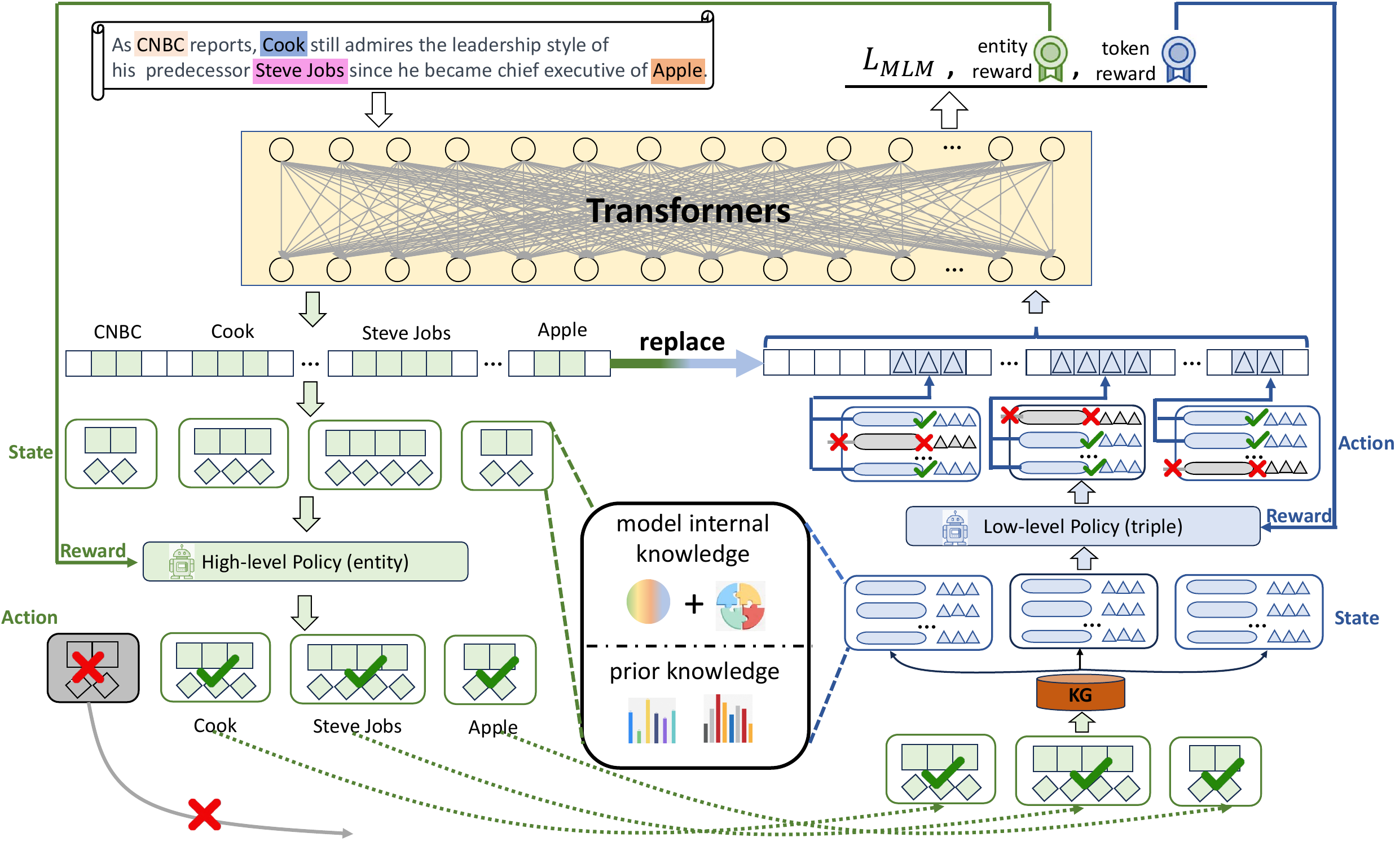}
\caption{The model architecture of KEHRL. The green part represents Reinforced Entity Position Detection. The blue part represents Reinforced Triple Semantic Refinement.}
\label{model_overview}
\end{figure*}

In this section, we introduce our model components in detail. An overview is shown in Figure \ref{model_overview}.
\subsection{Model Notations}
In the pre-training corpus, there are $N_{\text{total}}$ sentences, and each sentence consists of certain tokens $S_{i}=(t_{i1},t_{i2},\cdots ,t_{il_{i}})$.
Each sentence includes $N_{i}$ entities, and the sentence's entity collection can be denoted as $E_{i}=\{e_{i1},e_{i2},\cdots, e_{iN_{i}}\}$.
The $j$-th entity of the $i$-th sentence is connected with $M_{ij}$ triples in the KG; the entity's triple collection is denoted as $\{tri_{ij}^{1},tri_{ij}^{2},\cdots, tri_{ij}^{M_{ij}}\}$.
We further denote $d$ as the dimension of the model's hidden representations.

\subsection{Reinforced Entity Position Detection}
In this module, the policy learns to dynamically select the entity injection positions. The recognized entities in the sentence are regarded as the candidate pool.
The more semantically important entities are selected, the higher the reward the strategy owns.

\paragraph{State:}
The representation of each entity is obtained in a knowledge combination process (See Sec. \ref{enhanced_details}).
The representation of the $j$-th entity in the $i$-th sentence $e_{ij}$ is $H_{e_{ij}} \in \mathbb{R}^{d}$. We treat the concatenation of all the entity representations in the sample sentence as the state $s_{i}^{\text{high}}$ of RL:
\begin{equation}
\label{high_level_state}
    s_{i}^{\text{high}} = \{H_{e_{i1}} \; || \; H_{e_{i2}} \; || \; \cdots \; || \; H_{e_{iN_{i}}}\}
\end{equation}
where ``$||$'' denotes the operation of concatenation.

\paragraph{Policy:}
The policy of high-level RL is a probability distribution to decide which entity is more informative to the sentence.
It leverages the current state to conduct related actions $a_{i}^{\text{high}}$. The policy $\pi_{\theta_{\text{high}}}$ is formulated as follows:
\begin{equation}
\label{high_level_policy}
    \pi_{\theta_{\text{high}}} (a_{i}^{\text{high}} | s_{i}^{\text{high}})  = P(a_{i}^{\text{high}} | s_{i}^{\text{high}})
\end{equation}
where $\theta_{\text{high}}$ represents the parameters of the policy.

\paragraph{Action:}
The action of high-level RL is to select the entities, and the action for the $i$-th sentence is a binary value vector:
\begin{equation}
\label{high_level_action}
    \begin{split}
        a_{i}^{\text{high}} &= \{0,1, \cdots,0\} \in \mathbb{R}^{N_{i}}, \\
        a_{i}^{\text{high}} &\sim \pi_{\theta_{\text{high}}} (a_{i}^{\text{high}} | s_{i}^{\text{high}})
    \end{split}
\end{equation}
where ``1'' and ``0'' in the vector denote whether to select the entity or not.
``$\sim$'' means that the former is sampled from the distribution of the latter.
Note that the low-level RL process for dynamic triple selection is only triggered when the high-level action for selecting entity injection positions is performed.

\paragraph{Reward:}
Our work treats the entity-grained task performance as the high-level reward instead of the intermediate reward. This depends on whether the correct entity is predicted in the MLM task. The reward can be formulated as follows:
\begin{equation}
\label{high_level_reward}
    r_{ij}^{\text{high}}=\begin{cases}
                      1 & \text{if } \hat{e}_{ij} = e_{ij}, \\
                      0 & \text{if } \hat{e}_{ij} \ne e_{ij}.
                \end{cases}
\end{equation}
where $\hat{e}_{ij}$ is the $j$-th entity predicted by the model for sentence $S_{i}$.
The accumulated reward for the sentence is computed by $R_{i}^{\text{high}} = \sum_{j=1}^{|Mask_{i}^{e}|} r_{ij}^{\text{high}}$, where $|Mask_{i}^{e}|$ is the number of masked entities.

\subsection{Reinforced Triple Semantic Refinement}
At the low-level RL, the policy prompts the model to choose more accurate triples with respect to the selected entities during pre-training and filters out inaccurate triples, rather than relying on soft attention weights \cite{DBLP:conf/acl/ZhangHLJSL19,DBLP:conf/emnlp/LinCCR19,DBLP:conf/emnlp/PetersNLSJSS19}.

\paragraph{State:}  
We obtain the representation of each relation triple $H_{\text{tri}_{ij}^{p}} \in \mathbb{R}^{d}$ by knowledge combination (See Sec.~\ref{enhanced_details}).
The concatenation of all triple representations retrieved based on the selected entities is treated as the state of low-level RL.
Hence, the state is represented as:
\begin{equation}
\label{low_level_state}
    s_{ij}^{\text{low}} = \{H_{\text{tri}_{ij}^{1}} \; || \; H_{\text{tri}_{ij}^{2}} \; || \; \cdots \; || \; H_{\text{tri}_{ij}^{M_{ij}}}\}
\end{equation}

\paragraph{Policy:} The policy of low-level RL determines a subset of related triples for a selected entity. 
It utilizes the representations of the selected entity's and all of the connected triples' to conduct the choosing operation. The policy is a distribution given the high-level action and the low-level states, i.e.,
\begin{equation}
\label{low_level_policy}
    \pi_{\theta_{\text{low}}} (a_{ij}^{\text{low}} | a_{i}^{\text{high}}; s_{ij}^{\text{low}}) = P(a_{ij}^{\text{low}} | a_{i}^{\text{high}}; s_{ij}^{\text{low}})
\end{equation}

\paragraph{Action:} The action of low-level RL is selecting unambiguous relation triples and forcing the removal of inaccurate relation triples in the current training iteration.
The action for the $j$-th entity in the $i$-th sentence is also a binary value vector to represent whether to choose the triple or not:
\begin{equation}
\label{low_level_action}
    \begin{split}
        a_{ij}^{\text{low}} = \{0,1, \cdots, 0\} \in \mathbb{R}^{M_{ij}}, \\
        a_{ij}^{\text{low}} \sim \pi_{\theta_{\text{low}}} (a_{ij}^{\text{low}} | a_{i}^{\text{high}}; s_{ij}^{\text{low}})
    \end{split}
\end{equation}

\paragraph{Reward:} The final MLM task results at the token level are regarded as the low-level reward, similar to the high-level reward process.
The low-level reward is the number of correctly predicted tokens in the MLM task, i.e.,
\begin{equation}
\label{low_level_reward}
    r_{iq}^{\text{low}} = \begin{cases}
                      1 & \text{if } \hat{t}_{iq} = t_{iq}, \\
                      0 & \text{if } \hat{t}_{iq} \ne t_{iq}.
                \end{cases}
\end{equation}
where $\hat{t}_{iq}$ is the $q$-th token predicted by the model for sentence $S_{i}$. 
The accumulated reward for the sentence is computed as $R_{i}^{\text{low}} = \sum_{q=1}^{|Mask_{i}^{t}|} r_{iq}^{\text{low}}$, where $|Mask_{i}^{t}|$ is the number of masked tokens.
To inject the relation triples into PLMs, we utilize the triple representations to replace the related original entity representations at the entity positions.

\subsection{Weighted Knowledge Combination}
\label{enhanced_details}
We leverage the weighted mix of two types of knowledge, including the model's internal knowledge (i.e., the context-aware representations) and prior knowledge, to further optimize the learning process.

\paragraph{Model's Internal Knowledge:} Different granularities of text representations are combined to participate in the RL procedure.
The entity's internal knowledge consists of the contexts and KGs.
The contextual information is the mean-pooling representation $h_{e_{ij}} \in \mathbb{R}^{d}$, which is extracted from the sentence embedding between the entity start and end positions.
The entity's KG information is the merged representations of all its connected triples, i.e.,
\begin{equation}
\label{current_entity_emb}
    \begin{split}
        H_{e_{ij}}^{\text{mod}} = h_{e_{ij}} + \sum_{p=1}^{M_{ij}} \alpha_{ip} h_{\text{tri}_{ij}^{p}}, \\
        \alpha_{ip} = \frac{h_{e_{ij}} \cdot h_{\text{tri}_{ij}^{p}}}{\sum_{p=1}^{M_{ij}} h_{e_{ij}} \cdot h_{\text{tri}_{ij}^{p}}}.
    \end{split}
\end{equation}
The triple's internal knowledge consists of global-triple and sub-triple information.
The global-triple is the specific pseudo sentence representation $h_{\text{tri}_{ij}^{p}} \in \mathbb{R}^{d}$. 
The pseudo sentence is the concatenation of the triples $\text{tri}_{ij}^{p} = \langle \text{head}, \text{rel}, \text{tail} \rangle$.
The sub-triple is the merging of all triple components; we compute the subject, relation, and object representations respectively, and $k \in \{\text{head}, \text{rel}, \text{tail}\}$, i.e.,
\begin{equation}
\label{current_triple_emb}
    \begin{split}
        H_{\text{tri}_{ij}^{p}}^{\text{mod}} &= h_{\text{tri}_{ij}^{p}} + \sum_{k=1}^{3} \beta_{ij}^{pk} h_{\text{tri}_{pk}^{ij}}, \\
        \beta_{ij}^{pk} &= \frac{h_{\text{tri}_{ij}^{p}} \cdot h_{\text{tri}_{pk}^{ij}}}{\sum_{k=1}^{3} h_{\text{tri}_{ij}^{p}} \cdot h_{\text{tri}_{pk}^{ij}}}.
    \end{split}
\end{equation}

\paragraph{Prior Knowledge:} To calibrate the learning direction and avoid distorted forward steps, we consider prior knowledge as part of the enhanced knowledge components pre-processed before the model training stage.
The entity's prior knowledge is the normalized appearance frequency of each entity relative to all entities in the training corpus. 
The triple's prior knowledge is the entity's connected (i.e., $1$-hop, $2$-hop, $\cdots$, $k$-hop) triples' normalized importance, calculated by semantic similarity:
\begin{equation}
\label{prior_entity_and_tri_emb}
    \begin{split}
        H_{e_{ij}}^{\text{pri}} = \text{softmax}\left(\frac{ C_{e_{ij}}}{\sum_{i=1}^{N_{\text{total}}} \sum_{j=1}^{N_{i}} C_{e_{ij}} }\right), \\
        H_{\text{tri}_{ij}^{p}}^{\text{pri}} = \text{softmax}(\text{sim}(h_{\text{tri}_{ij}^{p}}, h_{S_{i}})).
    \end{split}
\end{equation}
where $C_{e_{ij}}$ denotes the number of appearances of entity $e_{ij}$. ``$\text{sim}(\cdot,\cdot)$'' denotes the similarity using the cosine function. The sentence embedding $h_{S_{i}} \in \mathbb{R}^{d}$ is the sentence representation.

Next, we mix these two types of knowledge in a weighted operation:
\begin{equation}
\label{combined_cur_prior_entity_and_tri_emb}
    \begin{split}
        H_{e_{ij}} &= \lambda H_{e_{ij}}^{\text{mod}} + (1-\lambda) H_{e_{ij}}^{\text{pri}}, \\
        H_{\text{tri}_{ij}^{p}} &= \lambda H_{\text{tri}_{ij}^{p}}^{\text{mod}} + (1-\lambda) H_{\text{tri}_{ij}^{p}}^{\text{pri}}.
    \end{split}
\end{equation}
where $\lambda$ is the mixing degree controlling parameter.

\subsection{Hierarchical Learning Strategy}
Our total learning objectives are composed of three parts: the cross-entropy loss of the sentence MLM task, the high-level RL objective, and the low-level RL objective.
The high-level RL objective is to maximize the expected accumulated rewards in the entity-grained MLM task:
\begin{equation}
    J_{\theta_{\text{high}}} = \mathbb{E}_{s_{i}^{\text{high}}, a_{i}^{\text{high}}, r_{ij}^{\text{high}}} \sum_{j=1}^{|Mask_{i}^{e}|} r_{ij}^{\text{high}}
\end{equation}
The low-level RL objective is to maximize the expected accumulated rewards in the token-level MLM task:
\begin{equation}
    J_{\theta_{\text{low}}} = \mathbb{E}_{s_{ij}^{\text{low}}, a_{ij}^{\text{low}}, r_{iq}^{\text{low}}} \sum_{q=1}^{|Mask_{i}^{t}|} r_{iq}^{\text{low}}
\end{equation}
Note that there are no discretized time steps in the episode \cite{DBLP:books/lib/SuttonB98,lapan2018deep}.
We model the episode as the training iteration.
We treat the final task reward as the accumulated reward during the calculation stage.
To enable the model to converge faster and exhibit smaller variance, we exploit policy gradient methods \cite{DBLP:conf/nips/SuttonMSM99} via the REINFORCE algorithm \cite{DBLP:journals/ml/Williams92} with a baseline \cite{DBLP:conf/uai/WeaverT01} to optimize the RL objectives.
The gradients for the high-level and low-level policy are as follows:
\begin{equation}
\label{gradient_updata_function_high_and_low}
    \begin{split}
        \bigtriangledown_{\theta _{\text{high}}} J_{\theta _{\text{high}}} = \mathbb{E}_{_{s_{i}^{\text{high}},a_{i}^{\text{high}},r_{ij}^{\text{high}} }} [(R_{i}^{\text{high}}-R_{\text{base}_{i}}^{\text{high}})\\
                                                                        \bigtriangledown_{\theta _{\text{high}}}\pi_{\theta_{ \text{high}}} (a_{i}^{\text{high}} | s_{i}^{\text{high}})]\\
        \bigtriangledown_{\theta _{\text{low}}} J_{\theta _{\text{low}}} = \mathbb{E}_{s_{ij}^{\text{low}},a_{ij}^{\text{low}},r_{iq}^{\text{low}} } [(R_{i}^{\text{low}}-R_{\text{base}_{i}}^{\text{low}})\\
                                                                        \bigtriangledown_{\theta _{\text{low}}} \pi_{\theta_{ \text{low}}} (a_{ij}^{\text{low}} | a_{i}^{\text{high}};s_{ij}^{\text{low}})]
    \end{split}
\end{equation}
Thus, the total training objective is:
\begin{equation}
\label{total_loss}
\begin{aligned}
\mathcal{L}_{\text{total}} &= \omega_{1}\mathcal{L}_{\text{MLM}} +\omega_{2}\mathcal{L}_{\text{high}}+ \omega_{3} \mathcal{L}_{\text{low }}\\
                    &= \omega_{1}\mathcal{L}_{\text{MLM}} -\omega_{2}J_{\theta _{\text{high}}} -  \omega_{3}J_{\theta _{\text{low}}}
\end{aligned}
\end{equation}
where $\omega_{1}, \omega_{2}, \omega_{3}$ are hyperparameters.

\section{Experiments}

\subsection{Data and Baselines}

\noindent \textbf{(1) Pre-training Data.} We fetch the pre-training samples from a wealth of knowledge resources, i.e., the English Wikipedia (2020/03/01). 
We obtain the detected entities' description text and related relation triples from WikiData5M \cite{DBLP:journals/tacl/WangGZZLLT21} using the entity linking tool TAGME \cite{DBLP:conf/cikm/FerraginaS10}. We follow ERNIE \cite{DBLP:conf/acl/ZhangHLJSL19} to complete the additional data processing stages. Finally, we obtain the pre-training data with 26 million samples, 3,085,345 entities, and 822 relation types.

\noindent \textbf{(2) Downstream Data.} Our work is evaluated by the LAMA benchmark\footnote{\url{https://github.com/facebookresearch/LAMA/tree/main}} \cite{DBLP:conf/emnlp/PetroniRRLBWM19}. The four evaluation datasets of LAMA include approximately 2,550,000 sentences. Additionally, we introduce the Open Entity \cite{choi-etal-2018-ultra} with about 6,000 examples for the entity typing task, CoNLL2003 \cite{DBLP:conf/conll/SangM03} with about 22,000 examples for the named entity recognition task, and TACRED \cite{DBLP:conf/emnlp/ZhangZCAM17} with 106,000 examples for the relation extraction task.

\noindent \textbf{(3) Baselines.} \textbf{ERNIE} \cite{DBLP:conf/acl/ZhangHLJSL19}, \textbf{KnowBERT} \cite{DBLP:conf/emnlp/PetersNLSJSS19}, and \textbf{KALM} \cite{DBLP:journals/corr/abs-2210-04105} inject the retrieved relevant entity embeddings into the model by an integrated entity linker. \textbf{KEPLER} \cite{DBLP:journals/tacl/WangGZZLLT21} and \textbf{DKPLM} \cite{DBLP:conf/aaai/Zhang0HQTH022} encode the entity's embedding and jointly optimize the model with knowledge embedding and MLM objectives. \textbf{GREASELM} \cite{DBLP:journals/corr/abs-2201-08860} fuses the graph structure and language context representation to encourage them to perform well on textual narratives tasks. \textbf{KP-PLM} \cite{DBLP:conf/emnlp/WangHQSWLG22} is trained with multiple transformed knowledge sub-graph prompts.

\subsection{Experiments Settings}

During the pre-training stage, we utilize the RoBERTa base model as the backbone. We choose AdamW \cite{DBLP:journals/corr/abs-1711-05101} as the optimizer with a learning rate of 5e-6 and a weight decay of 1e-5. The batch size is 184, and the model is trained for 5 epochs. The learning rate for the pre-training stage is set to 4e-5. The maximum hop number of the prior knowledge's entity connected triples $k$ is set to 3. In KEHRL, we fix the number of entities per sentence to 5 and the number of triples per entity to 7. Sentences with fewer than 5 entities randomly select entities from the existing ones to fill the total entity count. This rule also applies to triples. The maximum length of the concatenated triples $l_{\text{tri}}$ is 15. The ratio parameter $\lambda$ controlling the internal and prior knowledge is 0.5. The proportion parameters $\omega$ of the total loss $\mathcal{L}_{\text{total}}$ are set to \{0.3, 0.35, 0.35\}. We run our pre-training stage on 8 NVIDIA A100 80G GPUs for 1 day.\footnote{The source code and data can be available at \url{https://github.com/MatNLP/KEHRL}}

\begin{table*}[!htb]
\centering
\resizebox{\linewidth}{!}{
\begin{tabular}{c|cc|ccccccc}
\toprule
\multirow{2}{*}{Datasets$\downarrow$ \quad Models$\rightarrow$} & \multicolumn{2}{c|}{PLMs} & \multicolumn{7}{c}{KEPLMs}                                                                \\
                                                          & ELMo       & RoBERTa      & KEPLER & GREASELM & DKPLM & KP-PLM & \multicolumn{1}{c|}{KALM} & KEHRL         & $\Delta$ \\ \midrule
Google-RE                                                 & 2.2        & 5.3          & 7.3    & 10.6     & 10.8  & 11.0   & \multicolumn{1}{c|}{10.9} & \textbf{11.6} & +0.6     \\
UHN-Google-RE                                             & 2.3        & 2.2          & 4.1    & 5.0      & 5.4   & 5.6    & \multicolumn{1}{c|}{5.4}  & \textbf{5.9}  & +0.3     \\ \midrule
T-REx                                                     & 0.2        & 24.7         & 24.6   & 26.8     & 32.0  & 32.3   & \multicolumn{1}{c|}{31.1} & \textbf{34.9} & +2.6     \\
UHN-T-REx                                                 & 0.2        & 17.0         & 17.1   & 22.7     & 22.9  & 22.5   & \multicolumn{1}{c|}{23.1} & \textbf{27.5} & +4.4     \\ \bottomrule
\end{tabular}}
\caption{Experimental results of KEHRL and baselines on the LAMA benchmark in terms of Mean P@1 metric (\%). The t-tests demonstrate the improvements of KEHRL are statistically significant with $p$ <0.05.}
\label{general_results}
\end{table*}
\begin{table*}[tb]
\centering
\resizebox{\linewidth}{!}{
\begin{tabular}{c|ccc|ccc|ccc}
\toprule
\multirow{2}{*}{ Models$\downarrow$ \quad Datasets$\rightarrow$ } & \multicolumn{3}{c|}{Open Entity}                                                                                & \multicolumn{3}{c|}{CoNLL2003}                 & \multicolumn{3}{c}{TACRED}                                                                                     \\ \cmidrule{2-10} 
                       & Precision                      & Recall                         & F1                                           & Precision     & Recall        & F1            & Precision                      & Recall                         & F1                                           \\ \midrule
BERT                   & 76.4                           & 72.0                           & 73.6                                         & 91.6          & 93            & 92.4          & 67.2                           & 64.8                           & 66.0                                         \\
RoBERTa                & 77.4                           & 73.6                           & 75.4                                         & 90.9          & 94.4          & 92.6          & 70.8                           & 69.6                           & 70.2                                         \\ \midrule
ERNIE$_{BERT}$         & 78.4                           & 72.9                           & 75.6                                         & 89.5          & 94.2          & 91.8          & 70.0                           & 66.1                           & 68.1                                         \\
KnowBERT$_{BERT}$      & 77.9                           & 71.2                           & 74.4                                         & 91.2          & 92.8          & 92.0          & 71.6                           & 71.5                           & 71.5                                         \\
GREASELM               & 80.1                           & 74.8                           & 77.4                                         & 91.6          & 94.0          & 92.8          & 73.9                           & 73.2                           & 73.5                                         \\
DKPLM                  & 79.2                           & \textbf{75.9} & 77.5                                         & 92.5          & 93.7          & 93.1          & 72.6                           & 73.5                           & 73.1                                         \\
KP-PLM                 & 80.8                           & 75.1                           & 77.8                                         & 92.7          & 93.9          & 93.3          & 72.6                           & 73.7                           & 73.2                                         \\
KALM                   & 82.5                           & 75.2                           & 78.7                                         & 92.3          & 93.5          & 92.9          & 74.7                           & 73.8                           & 74.2                                         \\ \midrule
KEHRL                  & \textbf{89.3} & 75.6                           & \textbf{81.9$_{(\pm 0.7)}$} & \textbf{92.8} & \textbf{94.6} & \textbf{93.7$_{(\pm 0.3)}$} & \textbf{78.0} & \textbf{74.1} & \textbf{76.0$_{(\pm 0.4)}$} \\ \bottomrule
\end{tabular}}
\caption{The experimental results (\%) on downstream knowledge-intensive tasks.}
\label{knowledge_intenseve_results}
\end{table*}

\subsection{General Experimental Results}

\paragraph{Zero-shot Knowledge Probing Tasks:}
We evaluate our model on the knowledge probing benchmark LAMA \cite{DBLP:conf/emnlp/PetroniRRLBWM19} using the metric of macro-averaged mean precision (Mean P@1). As shown in Table \ref{general_results}, (1) KEHRL outperforms general PLMs and recent KEPLMs, owing to the RL-based knowledge selection. (2) The average improvements on T-REx related datasets are larger than those on Google-RE related datasets (3.5 vs. 0.5), demonstrating KEHRL's ability to probe factual knowledge in complex prediction scenarios. (3) The performance of KEHRL on four LAMA datasets is 2.1 points higher on average than the recent prevalent prompt-based model KP-PLM \cite{DBLP:conf/emnlp/WangHQSWLG22}.
We conjecture that KP-PLM focuses on prompt construction and pays less attention to filtering out inaccurate knowledge, indicating that our RL technique refines knowledge injection delicately and boosts performance.

\paragraph{Knowledge-intensive NLP Tasks via Fine-tuning:}
We validate our model on three knowledge-intensive tasks to verify its knowledge learning quality. The details are as follows:

In the entity linking task, we observe that (1) KEHRL achieves the best results in terms of the F1 metric and outperforms the strongest baseline KALM \cite{DBLP:journals/corr/abs-2210-04105} by 3.8 points (from 78.1 to 81.9) because KALM integrates abundant knowledge without meticulous detection of inaccurate samples. (2) Compared with the baselines, KEHRL attains the highest precision score of 89.3. These results confirm that the tailored knowledge injection mechanism of KEHRL effectively incorporates knowledge to enhance performance.
In the named entity recognition task, KEHRL achieves the best F1 score of 93.7 (+0.4) on this knowledge-intensive task, illustrating that our model has accurate representational ability in entity-aware scenarios, assisted by the knowledge refinement of RL.
In the relation extraction (RE) task, we fine-tune our model using the training set and test the model's relation extraction ability. The model yields the highest scores across all three metrics (+3.3 in Precision, +0.3 in Recall, and +1.8 in F1), further indicating KEHRL's accurate representations of entities and relations, due to the judicious knowledge injection.

\section{Detailed Analysis of KEHRL}

\subsection{Ablation Study} 
To evaluate the effectiveness of each important module in our model, we conduct an ablation study on the Open Entity \cite{choi-etal-2018-ultra} and TACRED \cite{DBLP:conf/emnlp/ZhangZCAM17} datasets. The results are as follows:
\begin{enumerate}
    \item When Reinforced Triple Semantic Refinement is removed, all related triples are injected into the model without meticulous refining, causing a performance drop of approximately 3.5 and 0.8 in F1 on the two datasets, respectively.
    \item Without Reinforced Entity Position Detection, knowledge is injected at all entity positions without essential position detection, which may introduce irrelevant knowledge, leading to a significant decline in model performance (81.9 $\rightarrow$ 74.2, and 76.0 $\rightarrow$ 71.8 in F1).
    \item Removing the weighted mixed knowledge and retaining only internal knowledge while ignoring prior knowledge results in a decrease of 1.4 and 0.5 points in F1.
\end{enumerate}
The ablation experiments indicate that each module contributes significantly to the model's performance on these tasks.

\begin{table}[tb]
\centering
\resizebox{\linewidth}{!}{
\begin{tabular}{lcc}
\toprule
\multicolumn{1}{l}{Methods}      & Open Entity & TACRED \\ \midrule
KEHRL                             & 81.9        & 76.0   \\
- Trip. Sema. Refi.       & 78.4        & 75.2   \\
- Enti. Posi. Dete. & 74.2        & 71.8   \\
- Weig. Mixed Knowledge       & 80.5        & 75.5   \\ \bottomrule
\end{tabular}}
\caption{Ablation study of KEHRL on Open Entity and TACRED in terms of F1 metric. ``-'' means removing the module.}
\label{ablation_study}
\end{table}

\begin{figure*}[tb]
\centering
\includegraphics[width=0.95\textwidth]{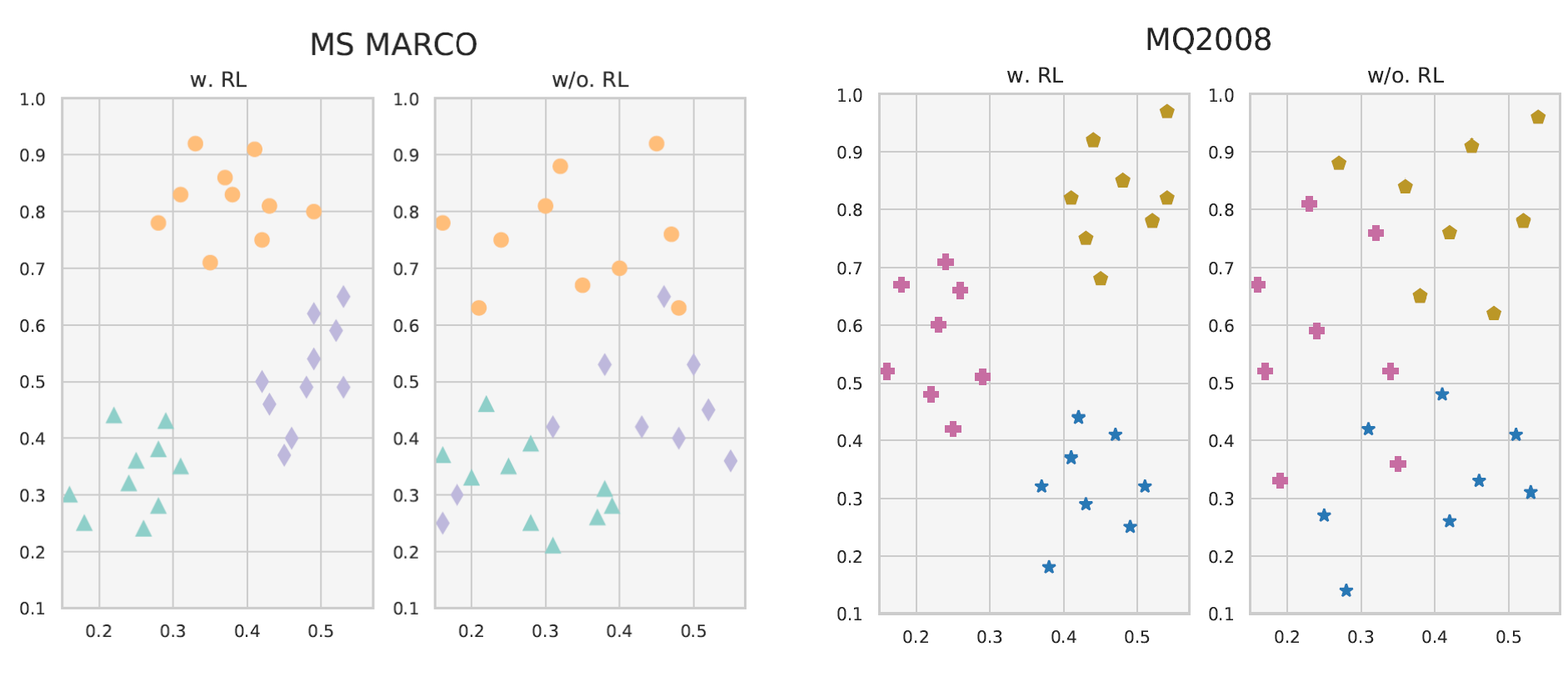} 
\vspace{-0.3cm}
\caption{The influence of Reinforcement Learning on MS MARCO and MQ2008.}
\label{influence_reinforcement_learning}
\end{figure*}

\subsection{The Influence of RL}

To probe the value of our customized knowledge injection operation, we compare our RL-refined model with a naive model that injects knowledge at all entity positions on QA tasks, including MS MARCO \cite{DBLP:conf/nips/NguyenRSGTMD16} and MQ2008 \cite{DBLP:journals/corr/QinL13}. We randomly sample three queries' top-10 related passages from MS MARCO and three queries' 8 related passages from MQ2008\footnote{In MQ2008, each query only contains 8 related passages.}, then feed these passages to the two different models.

In Figure \ref{influence_reinforcement_learning}, we visualize the passage representations after t-SNE \cite{van2008visualizing} dimensional reduction. The first and third figures represent passages from our meticulously RL-refined model, while the second and fourth figures represent passages from the model with naive knowledge injection at all entity positions. The closely clustered representations of our RL model indicate its capability to select accurate and informative knowledge. In contrast, the sparse distribution of the naive model's representations suggests less accurate knowledge selection.

\begin{figure*}[tb]
\centering
\includegraphics[width=0.9\textwidth]{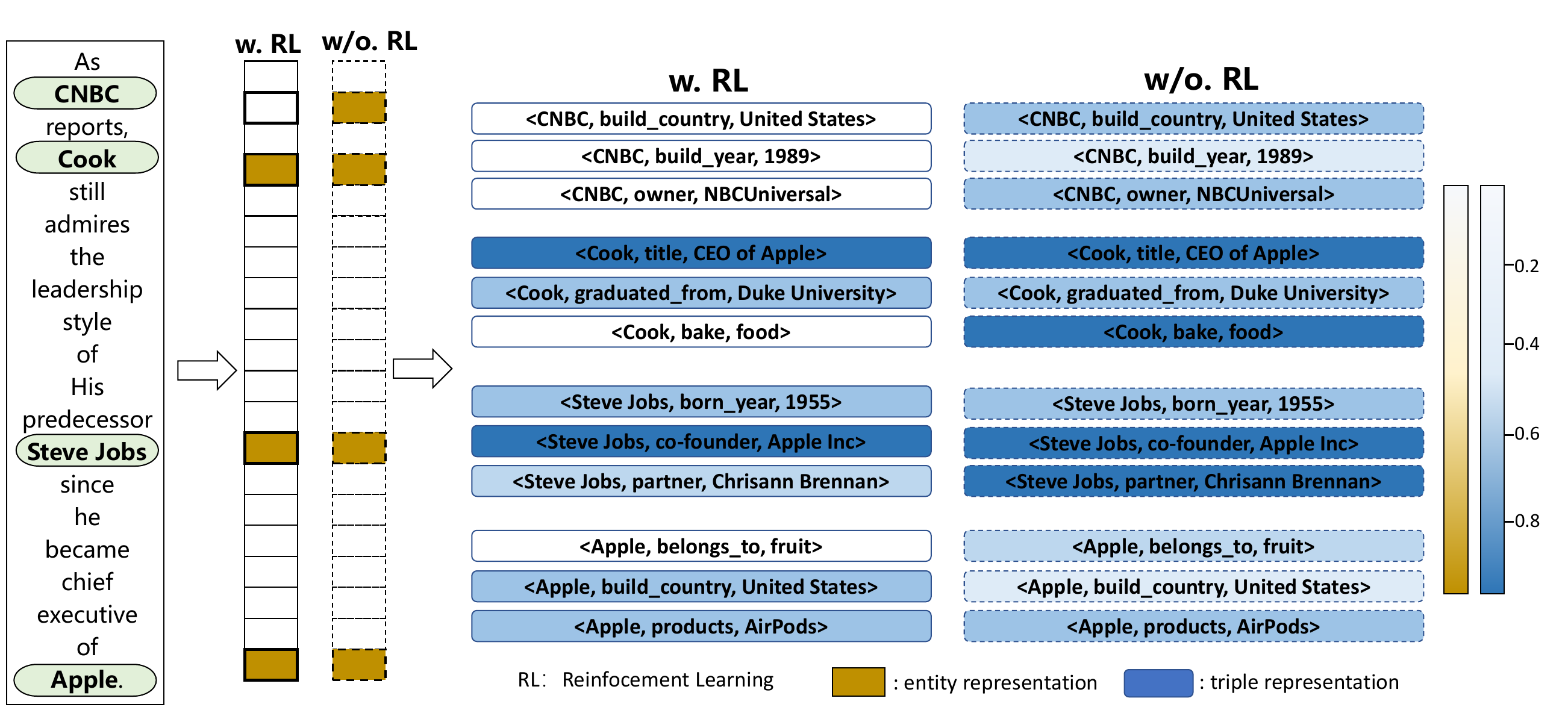}
\caption{The heat map of the entity and triple representations of KEHRL and the model without RL. The solid line produced by KEHRL and the dotted line is the model without RL.}
\label{heat_map}
\end{figure*}

A case study further highlights our model's effectiveness. As shown in Figure \ref{heat_map}, KEHRL correctly identifies meaningful entities and relation triples, allocating less attention to less significant entities such as ``CNBC'' and its related triples, thereby demonstrating the precision of our RL-based knowledge injection.

\subsection{The Influence of Weighted Mixed Knowledge} 

\begin{table}[tb]
\centering
\resizebox{\linewidth}{!}{
\begin{tabular}{lcc}
\toprule
Methods & Open Entity & TACRED     \\ \midrule
KEHRL                                                          & 81.9        & 76.0       \\ \midrule
No Prior                                                       & 80.5        & 75.5       \\
Element Type Prior                                            & 79.3        & 73.4       \\
Model Structure Prior                                          & {\ul 82.0}  & {\ul 76.3} \\
Data Augmentation Prior                                        & 81.7        & 75.9       \\ \bottomrule
\end{tabular}
}
\caption{The comparison of different prior knowledge combination methods.}
\vspace{-0.6cm}
\label{comparison_of_prior_knowledge}
\end{table}

We explore the impact of different prior knowledge strategies on our model:
\begin{itemize}
    \item \textbf{Element Type Prior:} Projects entity and relationship information into vectors for integration into the model.
    \item \textbf{Model Structure Prior:} Incorporates an additional model to preprocess training samples and generate representations for entities and triples.
    \item \textbf{Data Augmentation Prior:} Applies Easy Data Augmentation (EDA) techniques \cite{DBLP:conf/emnlp/WeiZ19} to samples and pseudo triple sentences, considering the augmented representations as prior knowledge.
\end{itemize}

According to the results presented in Table \ref{comparison_of_prior_knowledge}, both the No Prior and Element Type Prior strategies yield lower scores on the two datasets when compared to KEHRL. The Model Structure Prior achieves the best performance, surpassing that of KEHRL. However, this approach introduces an additional computational burden due to the extra model required. The Data Augmentation Prior performs marginally below KEHRL and entails increased computational costs associated with the data augmentation process. Consequently, our approach of leveraging linguistic prior knowledge proves to be the most cost-effective alternative.

\section{Related Works}

\subsection{Knowledge-Enhanced Pre-trained Language Models (KEPLMs)}

KEPLMs leverage external knowledge from Knowledge Graphs (KGs) to enhance semantic representation capabilities. They can be categorized based on the type of knowledge used:

\begin{enumerate}
    \item \textbf{Structured Knowledge:} Works by \citet{DBLP:conf/emnlp/ZhangDWWWLHLH22, DBLP:conf/emnlp/ZhangX0DCQCHQ23,DBLP:journals/aiopen/SuHZLLLZS21,DBLP:conf/coling/SunSQGHHZ20,DBLP:conf/emnlp/JiKHWZH20,DBLP:conf/emnlp/LinCCR19} augment models with sub-graphs from KGs by collecting multi-hop triples, learning nuanced semantics through graph neural networks and attention mechanisms.
    
    \item \textbf{Unstructured Knowledge:} \citet{DBLP:conf/acl/00020FYWX0022,DBLP:conf/ijcai/ChenLXYZ022} employ dictionary descriptions of sentence components to bolster the models' information retention. For instance, RAG \cite{DBLP:conf/nips/LewisPPPKGKLYR020} retrieves top-k related text documents or chunks using the K-NN algorithm to enrich the training corpus.
    
    \item \textbf{Heterogeneous Knowledge:} \citet{DBLP:conf/acl/QinLT00JHS020} incorporate both entity and relation representations in the neighboring space for enhancement. K-Adapter \cite{DBLP:conf/acl/WangTDWHJCJZ21} integrates contextual relation semantics of entities into the model through a pluggable training strategy.
\end{enumerate}

However, previous approaches generally treat knowledge integration as two separate processes, not considering entity selection and triple refinement jointly.

\subsection{Hierarchical Reinforcement Learning (HRL)}

Hierarchical Reinforcement Learning (HRL) breaks down complex problems into manageable sub-tasks, each addressed independently.

\begin{enumerate}
    \item \textbf{Top-down HRL:} This approach uses a high-level policy to determine low-level settings. For instance, \citet{DBLP:conf/aaai/TakanobuZLH19} divided the relation extraction task into high-level relation detection and low-level entity extraction. In medical applications, \citet{DBLP:journals/bioinformatics/ZhongLCLPHPW22} designed a master model to activate symptom checkers and disease classifiers. \citet{DBLP:journals/taslp/RohmatillahC23} established a high-level domain at the start of a dialogue, with sub-polices controlling the subsequent conversation.

    \item \textbf{Bottom-up HRL:} This type focuses on low-level policies aiding high-level policy learning. HRL-Rec \cite{DBLP:conf/aaai/XieZW0L21} has a low-level agent for channel selection, which guides high-level item recommendations. VISA \cite{DBLP:journals/jmlr/JonssonB06} decomposes the value function and employs a Dynamic Bayesian Network to model relationships.
\end{enumerate}

\section{Conclusion}

In this paper, we introduce KEHRL, a pre-training framework utilizing Hierarchical Reinforcement Learning for natural language understanding. The Reinforced Entity Position Detection module selects knowledge injection positions intelligently, avoiding less meaningful ones. The Reinforced Triple Semantic Refinement filters out inaccuracies and focuses on relevant triples linked to the chosen entities from the preceding module. Extensive experiments verify the effectiveness of KEHRL on factual knowledge based task and knowledge-intensive language tasks.

\section*{Acknowledgements}
We would like to thank anonymous reviewers for their valuable comments. This work was supported in part by National Key R\&D Program of China (No. 2022ZD0120302) and Alibaba Group through Alibaba Research Intern Program.
\section*{Bibliographical References}\label{sec:reference}

\bibliographystyle{lrec-coling2024-natbib}
\bibliography{languageresource}

\begin{thebibliography}{64}
\expandafter\ifx\csname natexlab\endcsname\relax\def\natexlab#1{#1}\fi

\bibitem[{Bao et~al.(2020)Bao, Dong, Wei, Wang, Yang, Liu, Wang, Gao, Piao, Zhou, and Hon}]{DBLP:conf/icml/Bao0WW0L0GP0H20}
Hangbo Bao, Li~Dong, Furu Wei, Wenhui Wang, Nan Yang, Xiaodong Liu, Yu~Wang, Jianfeng Gao, Songhao Piao, Ming Zhou, and Hsiao{-}Wuen Hon. 2020.
\newblock \href {http://proceedings.mlr.press/v119/bao20a.html} {Unilmv2: Pseudo-masked language models for unified language model pre-training}.
\newblock In \emph{ICML}, volume 119, pages 642--652.

\bibitem[{Chen et~al.(2022)Chen, Li, Xu, Yan, Zhang, and Zhang}]{DBLP:conf/ijcai/ChenLXYZ022}
Qianglong Chen, Feng{-}Lin Li, Guohai Xu, Ming Yan, Ji~Zhang, and Yin Zhang. 2022.
\newblock \href {https://doi.org/10.24963/ijcai.2022/567} {Dictbert: Dictionary description knowledge enhanced language model pre-training via contrastive learning}.
\newblock In \emph{IJCAI}, pages 4086--4092.

\bibitem[{Cheng et~al.(2021)Cheng, Shen, Liu, He, Chen, and Gao}]{DBLP:conf/acl/0002SLHCG20}
Hao Cheng, Yelong Shen, Xiaodong Liu, Pengcheng He, Weizhu Chen, and Jianfeng Gao. 2021.
\newblock \href {https://doi.org/10.18653/v1/2021.acl-long.240} {Unitedqa: {A} hybrid approach for open domain question answering}.
\newblock In \emph{ACL}, pages 3080--3090.

\bibitem[{Choi et~al.(2018)Choi, Levy, Choi, and Zettlemoyer}]{choi-etal-2018-ultra}
Eunsol Choi, Omer Levy, Yejin Choi, and Luke Zettlemoyer. 2018.
\newblock \href {https://doi.org/10.18653/v1/P18-1009} {Ultra-fine entity typing}.
\newblock In \emph{ACL}, pages 87--96.

\bibitem[{Devlin et~al.(2019)Devlin, Chang, Lee, and Toutanova}]{DBLP:conf/naacl/DevlinCLT19}
Jacob Devlin, Ming{-}Wei Chang, Kenton Lee, and Kristina Toutanova. 2019.
\newblock \href {https://doi.org/10.18653/v1/n19-1423} {{BERT:} pre-training of deep bidirectional transformers for language understanding}.
\newblock In \emph{NAACL}, pages 4171--4186.

\bibitem[{Dong et~al.(2019)Dong, Yang, Wang, Wei, Liu, Wang, Gao, Zhou, and Hon}]{DBLP:conf/nips/00040WWLWGZH19}
Li~Dong, Nan Yang, Wenhui Wang, Furu Wei, Xiaodong Liu, Yu~Wang, Jianfeng Gao, Ming Zhou, and Hsiao{-}Wuen Hon. 2019.
\newblock \href {https://proceedings.neurips.cc/paper/2019/hash/c20bb2d9a50d5ac1f713f8b34d9aac5a-Abstract.html} {Unified language model pre-training for natural language understanding and generation}.
\newblock In \emph{NeurIPS}, pages 13042--13054.

\bibitem[{Feng et~al.(2023)Feng, Tan, Zhang, Lei, and Tsvetkov}]{DBLP:journals/corr/abs-2210-04105}
Shangbin Feng, Zhaoxuan Tan, Wenqian Zhang, Zhenyu Lei, and Yulia Tsvetkov. 2023.
\newblock \href {https://doi.org/10.48550/arXiv.2210.04105} {{KALM:} knowledge-aware integration of local, document, and global contexts for long document understanding}.

\bibitem[{Ferragina and Scaiella(2010)}]{DBLP:conf/cikm/FerraginaS10}
Paolo Ferragina and Ugo Scaiella. 2010.
\newblock \href {https://doi.org/10.1145/1871437.1871689} {{TAGME:} on-the-fly annotation of short text fragments (by wikipedia entities)}.
\newblock In \emph{CIKM}, pages 1625--1628.

\bibitem[{Guu et~al.(2020)Guu, Lee, Tung, Pasupat, and Chang}]{DBLP:conf/icml/GuuLTPC20}
Kelvin Guu, Kenton Lee, Zora Tung, Panupong Pasupat, and Ming{-}Wei Chang. 2020.
\newblock \href {http://proceedings.mlr.press/v119/guu20a.html} {Retrieval augmented language model pre-training}.
\newblock In \emph{ICML}, volume 119, pages 3929--3938.

\bibitem[{Heo et~al.(2022)Heo, Kim, Choi, and Zhang}]{DBLP:conf/acl/HeoKCZ22}
Yu{-}Jung Heo, Eun{-}Sol Kim, Woo~Suk Choi, and Byoung{-}Tak Zhang. 2022.
\newblock \href {https://doi.org/10.18653/v1/2022.acl-long.29} {Hypergraph transformer: Weakly-supervised multi-hop reasoning for knowledge-based visual question answering}.
\newblock In \emph{ACL}, pages 373--390.

\bibitem[{Ji et~al.(2020)Ji, Ke, Huang, Wei, Zhu, and Huang}]{DBLP:conf/emnlp/JiKHWZH20}
Haozhe Ji, Pei Ke, Shaohan Huang, Furu Wei, Xiaoyan Zhu, and Minlie Huang. 2020.
\newblock \href {https://doi.org/10.18653/v1/2020.emnlp-main.54} {Language generation with multi-hop reasoning on commonsense knowledge graph}.
\newblock In \emph{EMNLP}, pages 725--736.

\bibitem[{Jonsson and Barto(2006)}]{DBLP:journals/jmlr/JonssonB06}
Anders Jonsson and Andrew~G. Barto. 2006.
\newblock \href {http://jmlr.org/papers/v7/jonsson06a.html} {Causal graph based decomposition of factored mdps}.
\newblock \emph{J. Mach. Learn. Res.}, 7:2259--2301.

\bibitem[{Lapan(2018)}]{lapan2018deep}
Maxim Lapan. 2018.
\newblock \href {https://books.google.com/books?hl=zh-CN&lr=&id=xKdhDwAAQBAJ&oi=fnd&pg=PP1&ots=wUgiiqXj9F&sig=fdx6QV8ezzmtPj5Z_z2x474QoeA#v=onepage&q&f=false} {\emph{Deep Reinforcement Learning Hands-On: Apply modern RL methods, with deep Q-networks, value iteration, policy gradients, TRPO, AlphaGo Zero and more}}.
\newblock Packt Publishing Ltd.

\bibitem[{Lee et~al.(2022)Lee, Li, Dozat, Perot, Su, Hua, Ainslie, Wang, Fujii, and Pfister}]{DBLP:conf/acl/LeeLDPSHAWFP22}
Chen{-}Yu Lee, Chun{-}Liang Li, Timothy Dozat, Vincent Perot, Guolong Su, Nan Hua, Joshua Ainslie, Renshen Wang, Yasuhisa Fujii, and Tomas Pfister. 2022.
\newblock \href {https://doi.org/10.18653/v1/2022.acl-long.260} {Formnet: Structural encoding beyond sequential modeling in form document information extraction}.
\newblock In \emph{ACL}, pages 3735--3754.

\bibitem[{Lewis et~al.(2020)Lewis, Perez, Piktus, Petroni, Karpukhin, Goyal, K{\"{u}}ttler, Lewis, Yih, Rockt{\"{a}}schel, Riedel, and Kiela}]{DBLP:conf/nips/LewisPPPKGKLYR020}
Patrick S.~H. Lewis, Ethan Perez, Aleksandra Piktus, Fabio Petroni, Vladimir Karpukhin, Naman Goyal, Heinrich K{\"{u}}ttler, Mike Lewis, Wen{-}tau Yih, Tim Rockt{\"{a}}schel, Sebastian Riedel, and Douwe Kiela. 2020.
\newblock \href {https://proceedings.neurips.cc/paper/2020/hash/6b493230205f780e1bc26945df7481e5-Abstract.html} {Retrieval-augmented generation for knowledge-intensive {NLP} tasks}.
\newblock In \emph{NeurIPS}.

\bibitem[{Li et~al.(2020)Li, Zhou, He, Wang, Yang, and Li}]{DBLP:conf/emnlp/LiZHWYL20}
Bohan Li, Hao Zhou, Junxian He, Mingxuan Wang, Yiming Yang, and Lei Li. 2020.
\newblock \href {https://doi.org/10.18653/v1/2020.emnlp-main.733} {On the sentence embeddings from pre-trained language models}.
\newblock In \emph{EMNLP}, pages 9119--9130.

\bibitem[{Lin et~al.(2019)Lin, Chen, Chen, and Ren}]{DBLP:conf/emnlp/LinCCR19}
Bill~Yuchen Lin, Xinyue Chen, Jamin Chen, and Xiang Ren. 2019.
\newblock \href {https://doi.org/10.18653/v1/D19-1282} {Kagnet: Knowledge-aware graph networks for commonsense reasoning}.
\newblock In \emph{EMNLP}, pages 2829--2839.

\bibitem[{Liu et~al.(2023)Liu, Jin, Wang, Cheng, Dou, and Wen}]{DBLP:journals/corr/abs-2306-05212}
Jiongnan Liu, Jiajie Jin, Zihan Wang, Jiehan Cheng, Zhicheng Dou, and Ji{-}Rong Wen. 2023.
\newblock \href {https://doi.org/10.48550/arXiv.2306.05212} {{RETA-LLM:} {A} retrieval-augmented large language model toolkit}.
\newblock \emph{CoRR}, abs/2306.05212.

\bibitem[{Liu et~al.(2020)Liu, Zhou, Zhao, Wang, Ju, Deng, and Wang}]{DBLP:conf/aaai/LiuZ0WJD020}
Weijie Liu, Peng Zhou, Zhe Zhao, Zhiruo Wang, Qi~Ju, Haotang Deng, and Ping Wang. 2020.
\newblock \href {https://ojs.aaai.org/index.php/AAAI/article/view/5681} {{K-BERT:} enabling language representation with knowledge graph}.
\newblock In \emph{AAAI}, pages 2901--2908.

\bibitem[{Liu et~al.(2022)Liu, Zhao, Gui, and Jin}]{DBLP:conf/dasfaa/LiuZGJ22}
Xiaokai Liu, Feng Zhao, Xiangyu Gui, and Hai Jin. 2022.
\newblock \href {https://doi.org/10.1007/978-3-031-00123-9\_9} {Lekan: Extracting long-tail relations via layer-enhanced knowledge-aggregation networks}.
\newblock In \emph{DASFAA}, volume 13245, pages 122--136.

\bibitem[{Loshchilov and Hutter(2017)}]{DBLP:journals/corr/abs-1711-05101}
Ilya Loshchilov and Frank Hutter. 2017.
\newblock \href {http://arxiv.org/abs/1711.05101} {Fixing weight decay regularization in adam}.
\newblock \emph{CoRR}, abs/1711.05101.

\bibitem[{Luo et~al.(2023)Luo, Xu, Zhao, Geng, Tao, Ma, Lin, and Jiang}]{DBLP:journals/corr/abs-2305-04757}
Ziyang Luo, Can Xu, Pu~Zhao, Xiubo Geng, Chongyang Tao, Jing Ma, Qingwei Lin, and Daxin Jiang. 2023.
\newblock \href {https://doi.org/10.48550/arXiv.2305.04757} {Augmented large language models with parametric knowledge guiding}.
\newblock abs/2305.04757.

\bibitem[{Ma et~al.(2021)Ma, Gui, Li, Zhang, Huang, and Zhou}]{DBLP:conf/acl/MaGLZHZ20}
Ruotian Ma, Tao Gui, Linyang Li, Qi~Zhang, Xuanjing Huang, and Yaqian Zhou. 2021.
\newblock \href {https://doi.org/10.18653/v1/2021.acl-long.484} {{SENT:} sentence-level distant relation extraction via negative training}.
\newblock In \emph{ACL}, pages 6201--6213.

\bibitem[{Ma et~al.(2020)Ma, Cui, Si, Liu, Wang, and Hu}]{DBLP:conf/coling/MaCSLWH20}
Wentao Ma, Yiming Cui, Chenglei Si, Ting Liu, Shijin Wang, and Guoping Hu. 2020.
\newblock \href {https://doi.org/10.18653/v1/2020.coling-main.4} {Charbert: Character-aware pre-trained language model}.
\newblock In \emph{COLING}, pages 39--50.

\bibitem[{Nguyen et~al.(2016)Nguyen, Rosenberg, Song, Gao, Tiwary, Majumder, and Deng}]{DBLP:conf/nips/NguyenRSGTMD16}
Tri Nguyen, Mir Rosenberg, Xia Song, Jianfeng Gao, Saurabh Tiwary, Rangan Majumder, and Li~Deng. 2016.
\newblock \href {https://ceur-ws.org/Vol-1773/CoCoNIPS\_2016\_paper9.pdf} {{MS} {MARCO:} {A} human generated machine reading comprehension dataset}.
\newblock In \emph{NIPS}, volume 1773 of \emph{{CEUR} Workshop Proceedings}.

\bibitem[{Pan et~al.(2023)Pan, Lin, Ge, Zhu, Zhang, Wang, Qiao, and Li}]{retrieval_llm}
Junting Pan, Ziyi Lin, Yuying Ge, Xiatian Zhu, Renrui Zhang, Yi~Wang, Yu~Qiao, and Hongsheng Li. 2023.
\newblock \href {https://arxiv.org/abs/2306.11732} {Retrieving-to-answer: Zero-shot video question answering with frozen large language models}.
\newblock \emph{CoRR}, abs/2306.11732.

\bibitem[{Pappas and Androutsopoulos(2021)}]{DBLP:conf/acl/PappasA20}
Dimitris Pappas and Ion Androutsopoulos. 2021.
\newblock \href {https://doi.org/10.18653/v1/2021.acl-long.301} {A neural model for joint document and snippet ranking in question answering for large document collections}.
\newblock In \emph{ACL}, pages 3896--3907.

\bibitem[{Peng et~al.(2023)Peng, Galley, He, Cheng, Xie, Hu, Huang, Liden, Yu, Chen, and Gao}]{DBLP:journals/corr/abs-2302-12813}
Baolin Peng, Michel Galley, Pengcheng He, Hao Cheng, Yujia Xie, Yu~Hu, Qiuyuan Huang, Lars Liden, Zhou Yu, Weizhu Chen, and Jianfeng Gao. 2023.
\newblock \href {https://doi.org/10.48550/arXiv.2302.12813} {Check your facts and try again: Improving large language models with external knowledge and automated feedback}.
\newblock abs/2302.12813.

\bibitem[{Peters et~al.(2019)Peters, Neumann, IV, Schwartz, Joshi, Singh, and Smith}]{DBLP:conf/emnlp/PetersNLSJSS19}
Matthew~E. Peters, Mark Neumann, Robert L.~Logan IV, Roy Schwartz, Vidur Joshi, Sameer Singh, and Noah~A. Smith. 2019.
\newblock \href {https://doi.org/10.18653/v1/D19-1005} {Knowledge enhanced contextual word representations}.
\newblock In \emph{EMNLP}, pages 43--54.

\bibitem[{Petroni et~al.(2019)Petroni, Rockt{\"{a}}schel, Riedel, Lewis, Bakhtin, Wu, and Miller}]{DBLP:conf/emnlp/PetroniRRLBWM19}
Fabio Petroni, Tim Rockt{\"{a}}schel, Sebastian Riedel, Patrick S.~H. Lewis, Anton Bakhtin, Yuxiang Wu, and Alexander~H. Miller. 2019.
\newblock \href {https://doi.org/10.18653/v1/D19-1250} {Language models as knowledge bases?}
\newblock In \emph{EMNLP}, pages 2463--2473.

\bibitem[{Qi et~al.(2022)Qi, Wan, Du, and Chen}]{DBLP:conf/acl/QiWDC22}
Kunxun Qi, Hai Wan, Jianfeng Du, and Haolan Chen. 2022.
\newblock \href {https://doi.org/10.18653/v1/2022.acl-long.134} {Enhancing cross-lingual natural language inference by prompt-learning from cross-lingual templates}.
\newblock In \emph{ACL}, pages 1910--1923.

\bibitem[{Qin and Liu(2013)}]{DBLP:journals/corr/QinL13}
Tao Qin and Tie{-}Yan Liu. 2013.
\newblock \href {http://arxiv.org/abs/1306.2597} {Introducing {LETOR} 4.0 datasets}.
\newblock \emph{CoRR}, abs/1306.2597.

\bibitem[{Qin et~al.(2021)Qin, Lin, Takanobu, Liu, Li, Ji, Huang, Sun, and Zhou}]{DBLP:conf/acl/QinLT00JHS020}
Yujia Qin, Yankai Lin, Ryuichi Takanobu, Zhiyuan Liu, Peng Li, Heng Ji, Minlie Huang, Maosong Sun, and Jie Zhou. 2021.
\newblock \href {https://doi.org/10.18653/v1/2021.acl-long.260} {{ERICA:} improving entity and relation understanding for pre-trained language models via contrastive learning}.
\newblock In \emph{ACL}, pages 3350--3363.

\bibitem[{Rohmatillah and Chien(2023)}]{DBLP:journals/taslp/RohmatillahC23}
Mahdin Rohmatillah and Jen{-}Tzung Chien. 2023.
\newblock \href {https://doi.org/10.1109/TASLP.2023.3235202} {Hierarchical reinforcement learning with guidance for multi-domain dialogue policy}.
\newblock \emph{{IEEE} {ACM} Trans}, 31:748--761.

\bibitem[{Saha et~al.(2020)Saha, Nie, and Bansal}]{DBLP:conf/emnlp/SahaNB20}
Swarnadeep Saha, Yixin Nie, and Mohit Bansal. 2020.
\newblock \href {https://doi.org/10.18653/v1/2020.emnlp-main.661} {Conjnli: Natural language inference over conjunctive sentences}.
\newblock In \emph{EMNLP}, pages 8240--8252.

\bibitem[{Sang and Meulder(2003)}]{DBLP:conf/conll/SangM03}
Erik F. Tjong~Kim Sang and Fien~De Meulder. 2003.
\newblock \href {https://aclanthology.org/W03-0419/} {Introduction to the conll-2003 shared task: Language-independent named entity recognition}.
\newblock In \emph{NAACL}, pages 142--147. {ACL}.

\bibitem[{Su et~al.(2021)Su, Han, Zhang, Lin, Li, Liu, Zhou, and Sun}]{DBLP:journals/aiopen/SuHZLLLZS21}
Yusheng Su, Xu~Han, Zhengyan Zhang, Yankai Lin, Peng Li, Zhiyuan Liu, Jie Zhou, and Maosong Sun. 2021.
\newblock \href {https://doi.org/10.1016/j.aiopen.2021.06.004} {Cokebert: Contextual knowledge selection and embedding towards enhanced pre-trained language models}.
\newblock \emph{{AI} Open}, 2:127--134.

\bibitem[{Sun et~al.(2020)Sun, Shao, Qiu, Guo, Hu, Huang, and Zhang}]{DBLP:conf/coling/SunSQGHHZ20}
Tianxiang Sun, Yunfan Shao, Xipeng Qiu, Qipeng Guo, Yaru Hu, Xuanjing Huang, and Zheng Zhang. 2020.
\newblock \href {https://doi.org/10.18653/v1/2020.coling-main.327} {Colake: Contextualized language and knowledge embedding}.
\newblock In \emph{COLING}, pages 3660--3670.

\bibitem[{Sutton and Barto(1998)}]{DBLP:books/lib/SuttonB98}
Richard~S. Sutton and Andrew~G. Barto. 1998.
\newblock \href {https://www.worldcat.org/oclc/37293240} {\emph{Reinforcement learning - an introduction}}.
\newblock Adaptive computation and machine learning. {MIT} Press.

\bibitem[{Sutton et~al.(1999)Sutton, McAllester, Singh, and Mansour}]{DBLP:conf/nips/SuttonMSM99}
Richard~S. Sutton, David~A. McAllester, Satinder Singh, and Yishay Mansour. 1999.
\newblock \href {http://papers.nips.cc/paper/1713-policy-gradient-methods-for-reinforcement-learning-with-function-approximation} {Policy gradient methods for reinforcement learning with function approximation}.
\newblock In \emph{NIPS}, pages 1057--1063.

\bibitem[{Takanobu et~al.(2019)Takanobu, Zhang, Liu, and Huang}]{DBLP:conf/aaai/TakanobuZLH19}
Ryuichi Takanobu, Tianyang Zhang, Jiexi Liu, and Minlie Huang. 2019.
\newblock \href {https://doi.org/10.1609/aaai.v33i01.33017072} {A hierarchical framework for relation extraction with reinforcement learning}.
\newblock In \emph{AAAI}, pages 7072--7079.

\bibitem[{van~der Maaten and Hinton(2008)}]{van2008visualizing}
Laurens van~der Maaten and Geoffrey Hinton. 2008.
\newblock Visualizing data using t-sne.
\newblock pages 2579--2605.

\bibitem[{Wang et~al.(2022)Wang, Huang, Qiu, Shi, Wang, Li, and Gao}]{DBLP:conf/emnlp/WangHQSWLG22}
Jianing Wang, Wenkang Huang, Minghui Qiu, Qiuhui Shi, Hongbin Wang, Xiang Li, and Ming Gao. 2022.
\newblock \href {https://aclanthology.org/2022.emnlp-main.207} {Knowledge prompting in pre-trained language model for natural language understanding}.
\newblock In \emph{EMNLP}, pages 3164--3177.

\bibitem[{Wang et~al.(2021{\natexlab{a}})Wang, Tang, Duan, Wei, Huang, Ji, Cao, Jiang, and Zhou}]{DBLP:conf/acl/WangTDWHJCJZ21}
Ruize Wang, Duyu Tang, Nan Duan, Zhongyu Wei, Xuanjing Huang, Jianshu Ji, Guihong Cao, Daxin Jiang, and Ming Zhou. 2021{\natexlab{a}}.
\newblock \href {https://doi.org/10.18653/v1/2021.findings-acl.121} {K-adapter: Infusing knowledge into pre-trained models with adapters}.
\newblock In \emph{Findings of ACL}, pages 1405--1418.

\bibitem[{Wang et~al.(2021{\natexlab{b}})Wang, Gao, Zhu, Zhang, Liu, Li, and Tang}]{DBLP:journals/tacl/WangGZZLLT21}
Xiaozhi Wang, Tianyu Gao, Zhaocheng Zhu, Zhengyan Zhang, Zhiyuan Liu, Juanzi Li, and Jian Tang. 2021{\natexlab{b}}.
\newblock \href {https://doi.org/10.1162/tacl\_a\_00360} {{KEPLER:} {A} unified model for knowledge embedding and pre-trained language representation}.
\newblock \emph{TACL}, 9:176--194.

\bibitem[{Weaver and Tao(2001)}]{DBLP:conf/uai/WeaverT01}
Lex Weaver and Nigel Tao. 2001.
\newblock \href {https://dslpitt.org/uai/displayArticleDetails.jsp?mmnu=1\&smnu=2\&article\_id=141\&proceeding\_id=17} {The optimal reward baseline for gradient-based reinforcement learning}.
\newblock In \emph{UAI}, pages 538--545.

\bibitem[{Wei and Zou(2019)}]{DBLP:conf/emnlp/WeiZ19}
Jason~W. Wei and Kai Zou. 2019.
\newblock \href {https://doi.org/10.18653/v1/D19-1670} {{EDA:} easy data augmentation techniques for boosting performance on text classification tasks}.
\newblock In \emph{EMNLP}, pages 6381--6387.

\bibitem[{Williams(1992)}]{DBLP:journals/ml/Williams92}
Ronald~J. Williams. 1992.
\newblock \href {https://doi.org/10.1007/BF00992696} {Simple statistical gradient-following algorithms for connectionist reinforcement learning}.
\newblock \emph{Mach. Learn.}, 8:229--256.

\bibitem[{Wu and He(2019)}]{DBLP:conf/cikm/WuH19a}
Shanchan Wu and Yifan He. 2019.
\newblock \href {https://doi.org/10.1145/3357384.3358119} {Enriching pre-trained language model with entity information for relation classification}.
\newblock In \emph{CIKM}, pages 2361--2364.

\bibitem[{Xie et~al.(2021)Xie, Zhang, Wang, Xia, and Lin}]{DBLP:conf/aaai/XieZW0L21}
Ruobing Xie, Shaoliang Zhang, Rui Wang, Feng Xia, and Leyu Lin. 2021.
\newblock \href {https://ojs.aaai.org/index.php/AAAI/article/view/16580} {Hierarchical reinforcement learning for integrated recommendation}.
\newblock In \emph{AAAI}, pages 4521--4528. {AAAI} Press.

\bibitem[{Yu et~al.(2022)Yu, Zhu, Fang, Yu, Wang, Xu, Zeng, and Jiang}]{DBLP:conf/acl/00020FYWX0022}
Wenhao Yu, Chenguang Zhu, Yuwei Fang, Donghan Yu, Shuohang Wang, Yichong Xu, Michael Zeng, and Meng Jiang. 2022.
\newblock \href {https://doi.org/10.18653/v1/2022.findings-acl.150} {Dict-bert: Enhancing language model pre-training with dictionary}.
\newblock In \emph{Findings of the ACL}, pages 1907--1918.

\bibitem[{Zhang et~al.(2023{\natexlab{a}})Zhang, Liu, Yuan, Fu, Chen, and Xiong}]{DBLP:journals/tkde/ZhangLYFCX23}
Denghui Zhang, Yanchi Liu, Zixuan Yuan, Yanjie Fu, Haifeng Chen, and Hui Xiong. 2023{\natexlab{a}}.
\newblock \href {https://doi.org/10.1109/TKDE.2022.3200921} {Multi-faceted knowledge-driven pre-training for product representation learning}.
\newblock \emph{TKDE}, 35(7):7239--7250.

\bibitem[{Zhang et~al.(2022{\natexlab{a}})Zhang, Zhang, Yu, Tang, Tang, Li, and Chen}]{DBLP:conf/acl/ZhangZY000C22}
Jing Zhang, Xiaokang Zhang, Jifan Yu, Jian Tang, Jie Tang, Cuiping Li, and Hong Chen. 2022{\natexlab{a}}.
\newblock \href {https://doi.org/10.18653/v1/2022.acl-long.396} {Subgraph retrieval enhanced model for multi-hop knowledge base question answering}.
\newblock In \emph{ACL}, pages 5773--5784.

\bibitem[{Zhang et~al.(2021{\natexlab{a}})Zhang, Deng, Cheng, Chen, Zhang, Zhang, and Chen}]{DBLP:conf/ijcai/ZhangDCCZZC21}
Ningyu Zhang, Shumin Deng, Xu~Cheng, Xi~Chen, Yichi Zhang, Wei Zhang, and Huajun Chen. 2021{\natexlab{a}}.
\newblock \href {https://doi.org/10.24963/ijcai.2021/552} {Drop redundant, shrink irrelevant: Selective knowledge injection for language pretraining}.
\newblock In \emph{IJCAI}, pages 4007--4014.

\bibitem[{Zhang et~al.(2023{\natexlab{b}})Zhang, Pan, Zhao, and Wang}]{DBLP:journals/corr/abs-2305-13669}
Shuo Zhang, Liangming Pan, Junzhou Zhao, and William~Yang Wang. 2023{\natexlab{b}}.
\newblock \href {https://doi.org/10.48550/arXiv.2305.13669} {Mitigating language model hallucination with interactive question-knowledge alignment}.
\newblock \emph{CoRR}, abs/2305.13669.

\bibitem[{Zhang et~al.(2021{\natexlab{b}})Zhang, Cai, Wang, Qiu, Yang, and He}]{DBLP:conf/acl/ZhangC0QYH20}
Taolin Zhang, Zerui Cai, Chengyu Wang, Minghui Qiu, Bite Yang, and Xiaofeng He. 2021{\natexlab{b}}.
\newblock \href {https://doi.org/10.18653/v1/2021.acl-long.457} {Smedbert: {A} knowledge-enhanced pre-trained language model with structured semantics for medical text mining}.
\newblock In \emph{ACL}, pages 5882--5893.

\bibitem[{Zhang et~al.(2022{\natexlab{b}})Zhang, Dong, Wang, Wang, Wang, Liu, Huang, Li, and He}]{DBLP:conf/emnlp/ZhangDWWWLHLH22}
Taolin Zhang, Junwei Dong, Jianing Wang, Chengyu Wang, Ang Wang, Yinghui Liu, Jun Huang, Yong Li, and Xiaofeng He. 2022{\natexlab{b}}.
\newblock \href {https://doi.org/10.18653/V1/2022.EMNLP-INDUSTRY.57} {Revisiting and advancing chinese natural language understanding with accelerated heterogeneous knowledge pre-training}.
\newblock In \emph{EMNLP}, pages 560--570.

\bibitem[{Zhang et~al.(2022{\natexlab{c}})Zhang, Wang, Hu, Qiu, Tang, He, and Huang}]{DBLP:conf/aaai/Zhang0HQTH022}
Taolin Zhang, Chengyu Wang, Nan Hu, Minghui Qiu, Chengguang Tang, Xiaofeng He, and Jun Huang. 2022{\natexlab{c}}.
\newblock \href {https://ojs.aaai.org/index.php/AAAI/article/view/21425} {{DKPLM:} decomposable knowledge-enhanced pre-trained language model for natural language understanding}.
\newblock In \emph{AAAI}, pages 11703--11711.

\bibitem[{Zhang et~al.(2023{\natexlab{c}})Zhang, Xu, Wang, Duan, Chen, Qiu, Cheng, He, and Qian}]{DBLP:conf/emnlp/ZhangX0DCQCHQ23}
Taolin Zhang, Ruyao Xu, Chengyu Wang, Zhongjie Duan, Cen Chen, Minghui Qiu, Dawei Cheng, Xiaofeng He, and Weining Qian. 2023{\natexlab{c}}.
\newblock \href {https://aclanthology.org/2023.emnlp-main.969} {Learning knowledge-enhanced contextual language representations for domain natural language understanding}.
\newblock In \emph{EMNLP}, pages 15663--15676.

\bibitem[{Zhang et~al.(2022{\natexlab{d}})Zhang, Bosselut, Yasunaga, Ren, Liang, Manning, and Leskovec}]{DBLP:journals/corr/abs-2201-08860}
Xikun Zhang, Antoine Bosselut, Michihiro Yasunaga, Hongyu Ren, Percy Liang, Christopher~D. Manning, and Jure Leskovec. 2022{\natexlab{d}}.
\newblock \href {http://arxiv.org/abs/2201.08860} {Greaselm: Graph reasoning enhanced language models for question answering}.

\bibitem[{Zhang et~al.(2017)Zhang, Zhong, Chen, Angeli, and Manning}]{DBLP:conf/emnlp/ZhangZCAM17}
Yuhao Zhang, Victor Zhong, Danqi Chen, Gabor Angeli, and Christopher~D. Manning. 2017.
\newblock \href {https://doi.org/10.18653/v1/d17-1004} {Position-aware attention and supervised data improve slot filling}.
\newblock In \emph{EMNLP}, pages 35--45.

\bibitem[{Zhang et~al.(2019)Zhang, Han, Liu, Jiang, Sun, and Liu}]{DBLP:conf/acl/ZhangHLJSL19}
Zhengyan Zhang, Xu~Han, Zhiyuan Liu, Xin Jiang, Maosong Sun, and Qun Liu. 2019.
\newblock \href {https://doi.org/10.18653/v1/p19-1139} {{ERNIE:} enhanced language representation with informative entities}.
\newblock In \emph{ACL}, pages 1441--1451.

\bibitem[{Zhao et~al.(2023)Zhao, Li, Joty, Qin, and Bing}]{DBLP:journals/corr/abs-2305-03268}
Ruochen Zhao, Xingxuan Li, Shafiq~R. Joty, Chengwei Qin, and Lidong Bing. 2023.
\newblock \href {https://doi.org/10.48550/arXiv.2305.03268} {Verify-and-edit: {A} knowledge-enhanced chain-of-thought framework}.
\newblock abs/2305.03268.

\bibitem[{Zhong et~al.(2022)Zhong, Liao, Chen, Liu, Peng, Huang, Peng, and Wei}]{DBLP:journals/bioinformatics/ZhongLCLPHPW22}
Cheng Zhong, Kangenbei Liao, Wei Chen, Qianlong Liu, Baolin Peng, Xuanjing Huang, Jiajie Peng, and Zhongyu Wei. 2022.
\newblock \href {https://doi.org/10.1093/bioinformatics/btac408} {Hierarchical reinforcement learning for automatic disease diagnosis}.
\newblock \emph{Bioinform.}, 38(16):3995--4001.

\end{thebibliography}


\end{document}